\documentclass[10pt,twocolumn,letterpaper]{article}

\usepackage{iccv}
\usepackage{times}
\usepackage{epsfig}
\usepackage{graphicx}
\usepackage{amsmath}
\usepackage{amssymb}

\usepackage[accsupp]{axessibility}
\usepackage{subfigure}
\usepackage{booktabs}
\usepackage{bm}
\usepackage[normalem]{ulem}
\usepackage{multirow}
\usepackage{makecell}
\usepackage{color, colortbl}
\usepackage[dvipsnames]{xcolor}
\definecolor{mypurple}{RGB}{160, 48, 160}
\definecolor{mygreen}{RGB}{84, 130, 53}
\definecolor{myorange}{RGB}{255, 210, 0}
\definecolor{myblue}{RGB}{0, 176, 240}
\definecolor{mybrown}{RGB}{190, 62, 19}
\definecolor{gray}{gray}{0.85}

\newcommand{\hyp}[0]{\mathbb{H}}

\newcommand{\norm}[1]{\left\lVert #1 \right\rVert}

\newcommand{\R}{\mathbb{R}}

\usepackage[pagebackref=true,breaklinks=true,letterpaper=true,colorlinks,bookmarks=false]{hyperref}

\iccvfinalcopy 

\def\iccvPaperID{3560} 

\begin{document}
\title{Hyperbolic Audio-visual Zero-shot Learning}

\author{Jie Hong$^{1,2}$, Zeeshan Hayder$^{2}$, Junlin Han$^{1,2}$, Pengfei Fang$^{3}$\thanks{Corresponding author},
\and
Mehrtash Harandi$^{4}$, Lars Petersson$^{2}$ \\
$^{1}$Australian National University, $^{2}$Data61-CSIRO,
$^{3}$Southeast University, $^{4}$Monash University \\
{\tt\small jie.hong@anu.edu.au},
{\tt\small zeeshan.hayder@data61.csiro.au},
{\tt\small junlinhcv@gmail.com}, \\
{\tt\small fangpengfei@seu.edu.cn},
{\tt\small mehrtash.harandi@monash.edu},
{\tt\small lars.petersson@data61.csiro.au}
}
\maketitle

\thispagestyle{empty}
\pagestyle{empty}

\begin{abstract}
Audio-visual zero-shot learning aims to classify samples consisting of a pair of corresponding audio and video sequences from classes that are not present during training. An analysis of the audio-visual data reveals a large degree of hyperbolicity, indicating the potential benefit of using a hyperbolic transformation to achieve curvature-aware geometric learning, with the aim of exploring more complex hierarchical data structures for this task. The proposed approach employs a novel loss function that incorporates cross-modality alignment between video and audio features in the hyperbolic space. Additionally, we explore the use of multiple adaptive curvatures for hyperbolic projections. The experimental results on this very challenging task demonstrate that our proposed hyperbolic approach for zero-shot learning outperforms the SOTA method on three datasets: VGGSound-GZSL, UCF-GZSL, and ActivityNet-GZSL achieving a harmonic mean (HM) improvement of around 3.0\%, 7.0\%, and 5.3\%, respectively. 
\end{abstract}

\section{Introduction}
Visual and audio signals frequently co-occur. For example, movies combine visual and auditory signals, providing an immersive experience to viewers. Humans perceive multiple sensory inputs jointly and make decisions accordingly. When a vehicle approaches from behind and honks, the driver sees the vehicle in the rear-view mirror, hears the sound, and decides to give way. The integration of joint visual and audio signals benefits numerous applications. For instance, acoustic features can help localize objects that emit sound in videos~\cite{lin2021exploring, zhou2021positive, wu2021binaural, qian2020multiple, ding2020self, senocak2018learning, pu2017audio}. 
Furthermore, the natural correlations between visual and audio signals provide strong supervision for learning video representations. As such, there has been growing interest in learning informative representations from audio signals for video classification~\cite{mittal2022learning, morgado2021robust, asano2020labelling, afouras2020self, cramer2019look, korbar2018cooperative, owens2016ambient}. The benefits of audio-visual multi-modality learning have also been demonstrated in other tasks, such as robotic navigation \cite{chen2020learning, gan2020look, dean2020see}, action recognition \cite{kazakos2019epic, gao2020listen}, highlight detection \cite{ye2021temporal, badamdorj2021joint}, violence detection \cite{wu2020not}, aerial scene recognition \cite{hu2020cross} and speech recognition \cite{xu2020discriminative, afouras2018deep}.

\begin{figure*}[th]
\centering
	\includegraphics[width=1.0\linewidth]{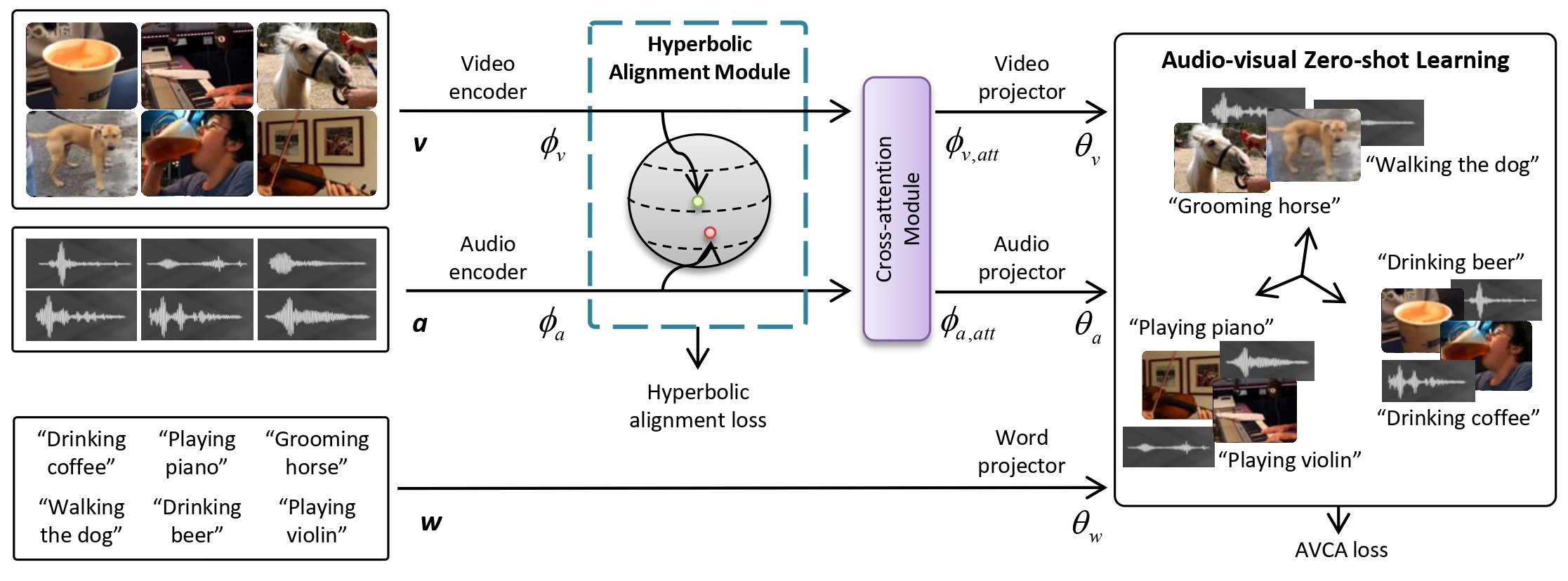}
 	\caption{We introduce the \textit{Hyperbolic Alignment Module}, represented by the block with the blue line. The input data features from each modality are encoded by two consecutive networks (encoder and projector). A cross-attention module in between is used to explore the natural correspondence between visual and audio features. Before the cross-attention module processes the features, our \textit{Hyperbolic Alignment Module} computes a hyperbolic alignment loss, which aims to explore more hierarchy in audio-visual data. For example, the model may discriminate embeddings of ``Playing piano'' and ``Walking the dog'' when it finds that these embeddings belong to different superclasses: ``Playing musical instruments'' and ``Walking/exercising/playing with animals''.}
 	\label{fig:baseline3}
\end{figure*}

Collecting vast amounts of audio-visual data needed for training deep neural networks is time-intensive, expensive, and in some applications, impractical.
Additionally, deep models may face difficulties in making accurate predictions when presented with objects from unseen classes in real-world scenarios. This is because they lack the necessary knowledge and context to make well-informed decisions about unfamiliar objects.
Low-shot audio-visual tasks \cite{parida2020coordinated} have emerged to address the problem of insufficient data, with an aim to generalize models effectively from seen to unseen data. These tasks include one-shot learning \cite{wang2021one}, few-shot learning \cite{majumder2022few}, and audio-visual zero-shot learning \cite{mercea2022audio, mercea2022temporal, mazumder2021avgzslnet, parida2020coordinated}. One-shot and few-shot audio-visual learning deals with classification from a few examples of unseen classes. However, audio-visual zero-shot classification poses a more challenging scenario as the model has no access to audio-visual data from unseen classes during training \cite{parida2020coordinated}. 

In this paper, we focus on audio-visual zero-shot learning. We investigate curvature-aware geometric learning for the audio-visual feature alignment. Our inspiration for using hyperbolic geometry comes from the following observations:

\begin{itemize}
\item \textbf{Data hierarchy.} Audio-visual datasets exhibit a hierarchy. For instance, in VGGSound \cite{chen2020vggsound}, all 309 classes can be categorized into 9 parent classes (or superclasses): ``people'', ``animals'', ``music'', ``sports'', ``nature'', ``vehicle'', ``home'', ``tools'' and ``others''. Similarly,  the dataset ActivityNet \cite{caba2015activitynet} provides a rich hierarchy with at least four levels. For example, the class ``Hand washing clothes" belongs to ``Laundry" (the 2$^{nd}$ level), ``Housework" (the 3$^{rd}$ level), and ``Household Activities" (the 4$^{th}$ level). Most existing audio-visual works have not adequately leveraged the hierarchical structure present in the audio-visual data.

\item \textbf{Hyperbolic geometric properties.} Hyperbolic methods have been shown to be effective in addressing low-shot visual problems \cite{khrulkov2020hyperbolic, liu2020hyperbolic,fang2021kernel,Fang2023_PoincareKernel,ma2022adaptive}. The learned hyperbolic feature embeddings have the ability to capture the hierarchical structure within the data. This is attributed to the tree-like nature
of the underlying space, as shown in \cite{guo2022co, chami2021horopca}. One of the benefits of using a hyperbolic space is that the hyperbolic space facilitates the distribution of embeddings in a tree-shaped structure since its volume expands exponentially.
\end{itemize}
Based on the observations above, we conjecture that the unique properties of hyperbolic spaces can be leveraged to capture the hierarchical structures in audio-visual data, leading to learning more discriminative embeddings for audio-visual samples. The current SOTA audio-visual zero-shot learning methods \cite{mercea2022audio, mercea2022temporal, mazumder2021avgzslnet, parida2020coordinated, zheng2023generative} operate in the non-curved Euclidean space without considering the data hierarchy. Therefore, there is a need for curvature-aware geometric solutions that can embed the data hierarchy to improve the performance of audio-visual zero-shot learning. The contributions of this work can be summarized as: 

\begin{itemize}
\item Our work provides a new perspective on using curved geometries for cross-modality, as shown in Figure~\ref{fig:baseline3}. We propose a hyperbolic alignment loss that learns features in curved space to improve audio-visual zero-shot learning. Specifically, we use alignment between visual and audio features in the hyperbolic space as an auxiliary learning method for feature fusion across modalities. To the best of our knowledge, we are the first to apply curvature-aware geometric solutions to this task.

\item Furthermore, we introduce various frameworks for using the hyperbolic embeddings: 1) Hyper-alignment, 2) Hyper-single, and 3) Hyper-multiple. The Hyper-alignment module maps audio and visual features from Euclidean to hyperbolic space with a fixed negative curvature and compares them using intra-modal similarities. Based on Hyper-alignment, Hyper-single adapts the curvature to the model for flexible data structure exploration. Hyper-multiple generates a set of adaptive curvatures for alignments, enabling more generic embeddings.

\item Extensive experiments demonstrate that, in most cases, the proposed modules outperform existing models. Moreover, using the $\delta_{rel}$ metric \cite{khrulkov2020hyperbolic, fournier2015computing}, we show that hyperbolic alignment enables the learned features to exhibit stronger hierarchical properties. Additionally, we observe from t-SNE \cite{van2008visualizing} visualizations that the audio-visual feature embeddings from different superclasses become more distinct.

\item Ablation studies are provided further to investigate the properties of our hyperbolic alignment module. In addition to the curvature-negative hyperbolic projection approach, we also test Euclidean and Spherical approaches, which have curvature-zero and curvature-positive properties, respectively. 
\end{itemize}

\section{Related Works}
\subsection{Audio-visual Zero-shot Learning}
The audio-visual zero-shot learning settings are first proposed in \cite{parida2020coordinated}, where the Coordinated Joint Multimodal Embedding (CJME) model is introduced to map video, audio, and text into the same feature space and compare them. The triplet loss was used to push video or audio features closer to their corresponding class features. Subsequently, Mazumder \etal \cite{mazumder2021avgzslnet} develop the Audio-Visual Generalized Zero-shot Learning Network (AVGZSLNet) to address audio-visual zero-shot learning, which includes a module that reconstructs text features from  visual and audio features. The Audio-Visual Cross-Attention (AVCA) framework \cite{mercea2022audio} is also designed to exchange information between video and audio representations, enabling informative representations help achieve state-of-the-art performance in audio-visual zero-shot classification. A similar design to AVCA for handling temporal data features is presented in \cite{mercea2022temporal}. 

\subsection{Hyperbolic Geometry}
Hyperbolic geometry has gained considerable interest in the machine learning community since its property of the negative curvature can encode the inherent hierarchical structure of the data. It benefits from a family of learning scenarios~\cite{Wei2022_hyperbolic_PAMI}, including sentiment analysis, recommendation systems, social network studies, etc. Ganea \etal first study the necessary to integrate hyperbolic geometry in the learning community due to its intriguing property to encode the hierarchical structure of the data and develop some necessary neural operators in the Poincar\'e ball~\cite{Ganea2018_HyperbolicNN_NIPS}. Its superior property is also leveraged to analyze and understand the irregular graph data~\cite{Chami2019_HyperbolicGCNN_NIPS}. Joint efforts are made to develop hyperbolic learning algorithms in the NLP and graph neural network field, \eg, hyperbolic attention network (HAN)~\cite{HAN_ICLR}, hyperbolic graph attention network (HGAN)~\cite{zhang2021hyperbolicGraphAtt} \etc. Khrulkov \etal analyze that the hierarchical structure also exists in the visual data and successfully integrates the hyperbolic space in the visual domain~\cite{khrulkov2020hyperbolic}. The concurrent work in \cite{liu2020hyperbolic} also shows that hyperbolic space is a good alternative for aligning the visual and text domains. The following works prove that hyperbolic geometry can benefit a series of visual tasks~\cite{fang2023hyperbolicCV}, like semantic segmentation~\cite{Atigh_2022_CVPR}, medical image recognition~\cite{Yu2022_MICCAI}, action recognition~\cite{Long_2020_CVPR}, anomaly recognition~\cite{hong2022curved} \etc. 

In this paper, we first learn hyperbolic embeddings for both visual and audio modalities. We then introduce a novel hyperbolic alignment loss that minimizes the differences between different modalities. This is different from existing approaches~\cite{mercea2022audio, mercea2022temporal, mazumder2021avgzslnet, parida2020coordinated, zheng2023generative} that focus solely on cross-modal correspondence between visual and audio streams. In contrast, our module considers cross-modal alignment in curvature-negative space (hyperbolic space) for mining hierarchical structures in audio-visual data.

\section{Preliminary}
To construct the hyperbolic alignment loss, we project the embedding point from Euclidean onto the hyperbolic tangent space. This involves two procedures: the hyperbolic projection and the logarithmic map.

\noindent \textbf{Hyperbolic projection.} In this paper, we utilize the Poincar'e ball \cite{nickel2017poincare, khrulkov2020hyperbolic} to model the hyperbolic space. The Poincar'e ball can be visualized as a ball space with a radius of $r = \sqrt{1/|c|}$, where $c<0$ represents the curvature. 
To project a point $\bm{x} \in \R^{n}$ from the Euclidean space onto the $n$-dimensional Poincar'e ball $\hyp_{c}^{n}$ with curvature $c$, we perform the hyperbolic projection as follows:  
\begin{equation}    \label{eqn:hyperbolic_projection}
\bm{x}_{H}= \Gamma_\hyp(\bm{x}) =
\begin{cases}
    \bm{x}   &  \text{if}~~\|\bm{x} \| \leq \frac{1}{|c|} \\
    \frac{1-\xi}{|c|} \frac{\bm{x}}{ \norm{\bm{x}}}
        &  \text{else}
\end{cases}                
\end{equation}
where the projected point in the Poincar'e ball is denoted by $\bm{x}_H \in \hyp_{c}^{n}$. To ensure numerical stability, a small value $\xi > 0$ is introduced. The addition of two points, $\bm{x}_H$ and $\bm{y}_H \in \hyp_{c}^{n}$, in the Poincar'e ball can be obtained via the M\"obius addition \cite{khrulkov2020hyperbolic} as follows: 
\begin{equation}
\resizebox{0.49\textwidth}{!}{
$
\bm{x}_H \oplus_{c} \bm{y}_H
= \frac{(1 + 2|c| \langle \bm{x}_H, \bm{y}_H\rangle + |c| \|\bm{y}_H\|^2)\bm{x}_H + (1-|c| \|\bm{x}_H \|^2)\bm{y}_H}{1 +2|c| \langle \bm{x}_H, \bm{y}_H \rangle + |c|^2 \|\bm{x}_H\|^2\|\bm{y}_H\|^2}
$
}
\end{equation}
where $\langle, \rangle$ is the inner product.

\noindent \textbf{Logarithmic map.}
The hyperbolic tangent space is an Euclidean space that locally approximates the hyperbolic space. 
By selecting a tangent point $\bm{z}_H \in \hyp_{c}^{n}$, we can generate a tangent plane $T_{z}\hyp_{c}^{n}$ at $\bm{z}_H$. The process of taking the logarithm is used to map a point $\bm{x}_H \in \hyp_{c}^{n}$ onto the tangent space $T_{z}\hyp_{c}^{n}$ of $\bm{z}_H$, which is given as follows:
\begin{equation} \label{eqn:tangent_projection}
\resizebox{0.49\textwidth}{!}{
$
\bm{x}_{Tg} = \frac{2}{\sqrt{|c|}\lambda_c(\bm{z}_H)}\mathrm{tanh}^{-1}(\sqrt{|c|}\| -\bm{z}_H \oplus_{c} \bm{x}_H \|) \frac{-\bm{z}_H \oplus_{c} \bm{x}_H}{\| -\bm{z}_H \oplus_{c} \bm{x}_H \|}
$
}
\end{equation}
where $\bm{x}_{Tg} \in T_{z}\hyp_{c}^{n}$ is the transformed point in the tangent space. In this paper, we consider the identity tangent space of the Poincar'e ball at the origin $\bm{0}_{H} \in \hyp_{c}^{n}$, so we have $\bm{z}_H = \bm{0}_{H}$.

\section{Hyperbolic Audio-visual Learning}
Audio-visual zero-shot learning aims to recognize audio-visual samples $\bm{u}=(\bm{v}_u, \bm{a}_u)$ from unseen classes during the evaluation stage \cite{parida2020coordinated}. In this task, the training data is restricted to the seen classes. In line with \cite{mercea2022audio}, we adopt the generalized setting where both seen and unseen class samples, $\bm{s}=(\bm{v}_s, \bm{a}_s)$ and $\bm{u}$, appear at the test stage. As illustrated in Figure~\ref{fig:baseline3}, the corresponding semantic words $\bm{w}$ are provided along with the audio-visual sample $(\bm{v}, \bm{a})$ to serve as textual labels.

\begin{figure*}[t]
\centering    
    \subfigure[Hyper-alignment]{\includegraphics[width=0.49\linewidth]{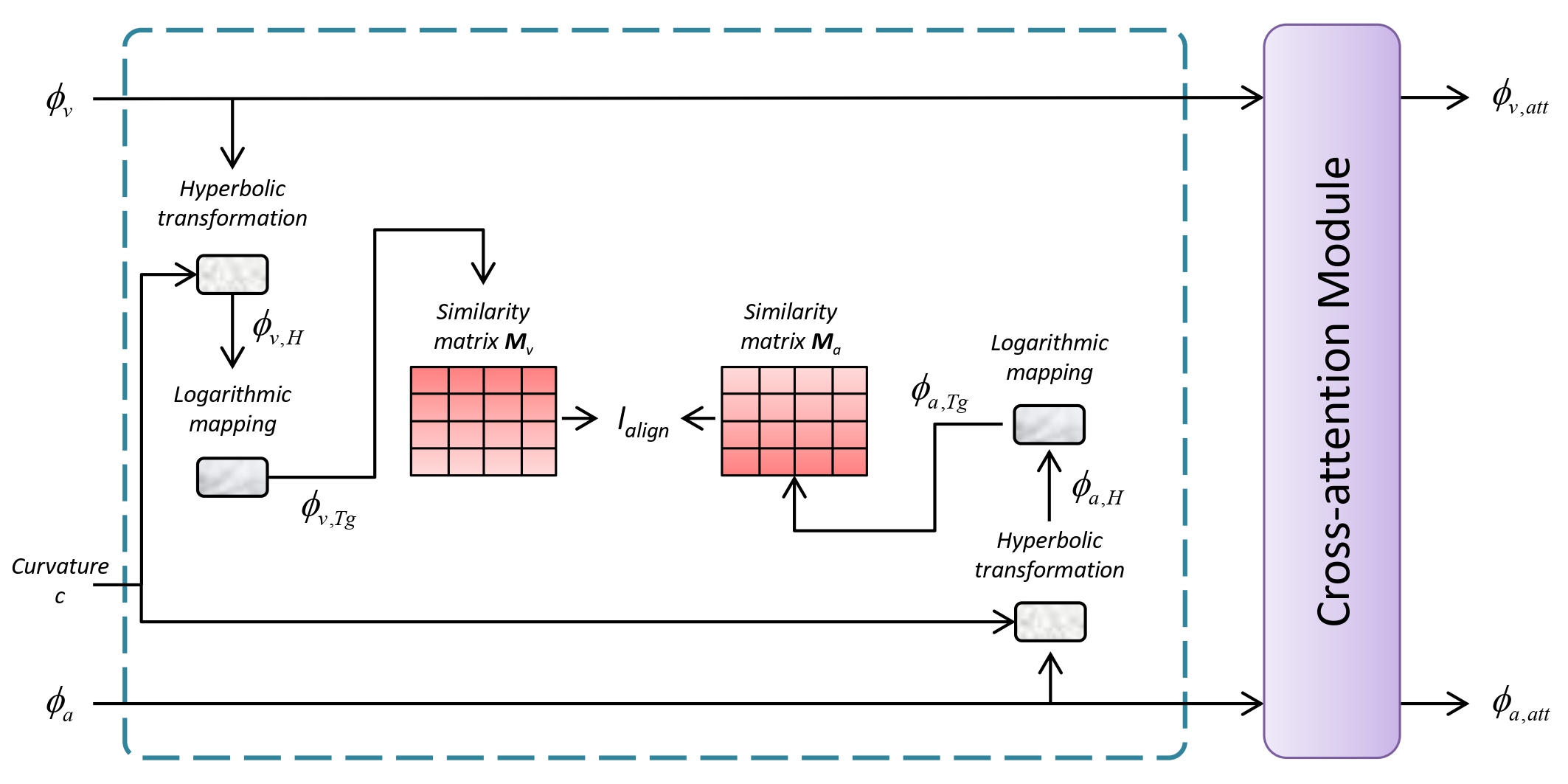}}
    \setcounter{subfigure}{2}
    \subfigure[Hyper-multiple]{\includegraphics[width=0.49\linewidth]{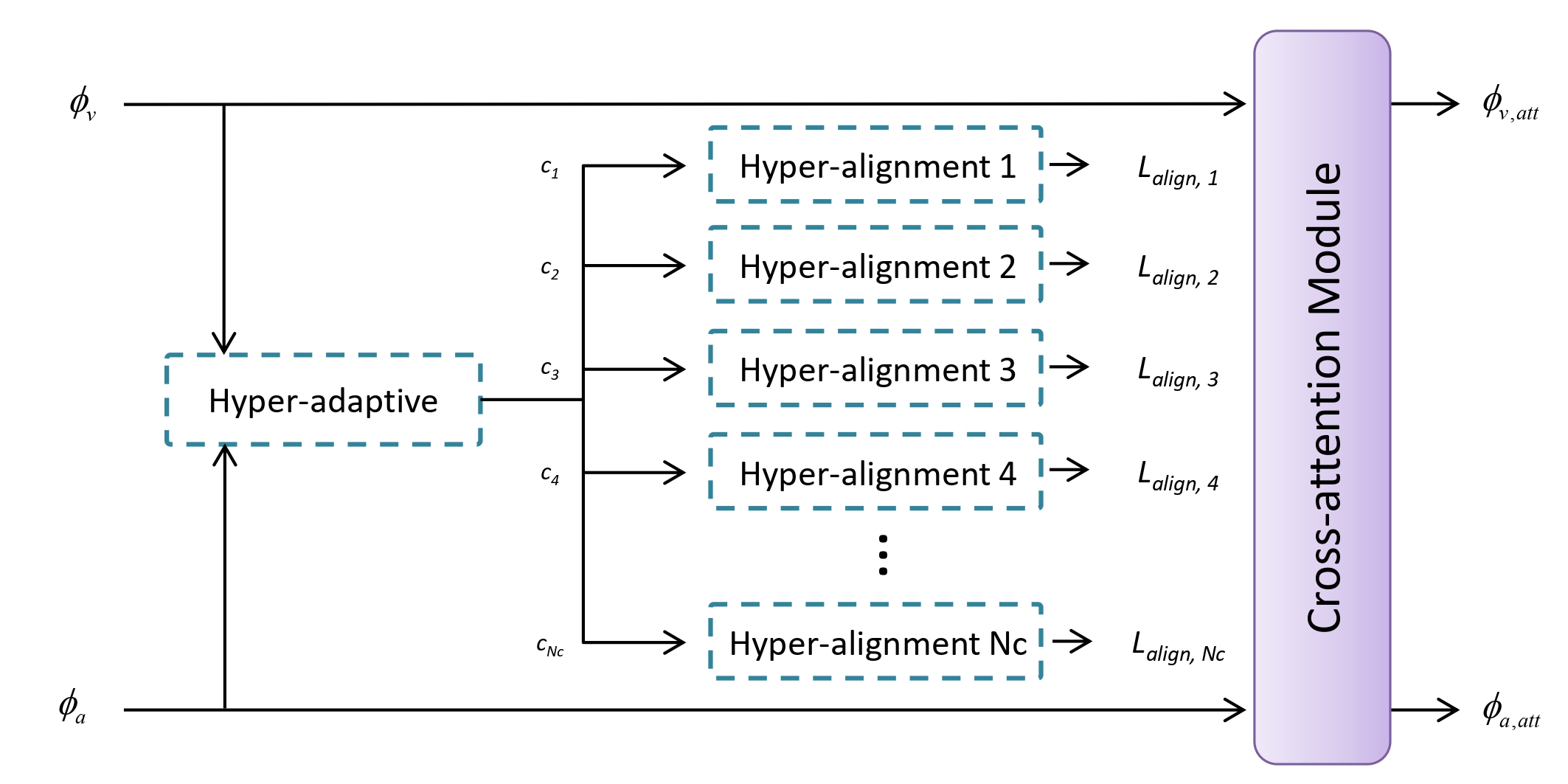}} \\
    \setcounter{subfigure}{1}
    \subfigure[Hyper-single]{\includegraphics[width=0.80\linewidth]{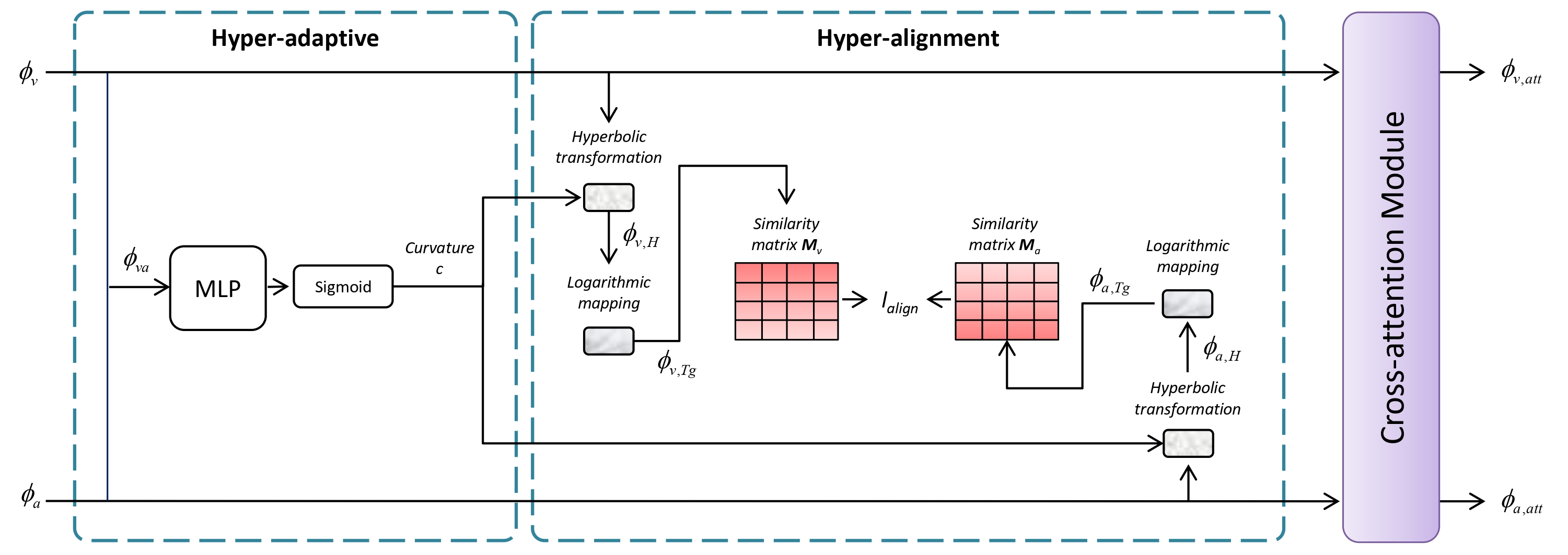}}
  
    \vspace{2mm}
 	\caption{The frameworks of the proposed \textit{Hyperbolic Alignment Module}: (a) \textit{Hyper-alignment}, which utilizes the hyperbolic space to learn feature alignment, (b) \textit{Hyper-single}, which automatically computes the curvature using the Hyper-adaptive module after concatenating visual and audio features as input, and (c) \textit{Hyper-multiple}, which performs alignment using multiple learnable curvatures. While Hyper-alignment uses a fixed curvature, Hyper-single and Hyper-multiple enable flexible exploration of intrinsic data structures by adapting the curvature. Hyper-multiple is particularly suited when diverse manifold structures coexist, improving the generality of the learned representations.}
 	\label{fig:hypermodel}
\end{figure*}

In this section, we first introduce the baseline AVCA \cite{mercea2022audio} and then provide details of our approach using a hyperbolic alignment loss for audio-visual zero-shot learning. As depicted in Figure~\ref{fig:baseline3}, we embed the network with a \textit{Hyperbolic Alignment Module} for computing the hyperbolic alignment loss. We propose three designs of this module: \textit{Hyper-alignment}, \textit{Hyper-single}, and \textit{Hyper-multiple}, as shown in Figure~\ref{fig:hypermodel} (a), (b) and (c). The subsequent subsections provide more details on each of the proposed designs.

\subsection{Baseline}
Our method is based on AVCA \cite{mercea2022audio}, which serves as the baseline. AVCA is a two-branch network that operates on synchronized visual and audio inputs, as shown in Figure~\ref{fig:baseline3}. These inputs, denoted by $\bm{v}, \bm{a} \in \R^{512}$ respectively, are represented by feature vectors extracted from pre-trained feature extractors, with both vectors having a dimension of $512$.
In addition to the video clip feature $\bm{v}$ and corresponding audio feature $\bm{a}$, the baseline model takes the text feature $\bm{w} \in \R^{512}$ as input. The features $\bm{\phi}_v, \bm{\phi}_a \in \R^{300}$ are obtained after passing $\bm{v}$ and $\bm{a}$ through the video and audio encoder, respectively. With $\bm{\phi}_v$ and $\bm{\phi}_a$, a cross-attention module calculates the cross-modality representations $\bm{\phi}_{v,att}, \bm{\phi}_{a,att} \in \R^{300}$. These features are then passed through the video and audio projectors to compute the latent embeddings $\bm{\theta}_v, \bm{\theta}_a \in \R^{64}$ in the common space for classifying audio-visual inputs $(\bm{v}, \bm{a})$. The text feature $\bm{w}$ is processed by the word projector to give $\bm{\theta}_w \in \R^{64}$ in the common space. The loss of AVCA is represented by $\mathcal{L}_{avca}$. 

\subsection{Hyper-alignment}
We propose a minimalist design called \textit{Hyper-alignment}, which learns feature alignment in a hyperbolic manifold using a fixed curvature, as illustrated in Figure~\ref{fig:hypermodel} (a). To construct the loss, both modality features $\bm{\phi}_v$ and $\bm{\phi}_a$ are first mapped to points $\bm{\phi}_{v,H}$ and $\bm{\phi}_{a,H}$ in the Poincar'e ball space using Eq.~(\ref{eqn:hyperbolic_projection}). Then, we obtain $\bm{\phi}_{v,Tg}$ and $\bm{\phi}_{a,Tg}$ in $T_{z}\hyp_{c}^{300}$ by projecting $\bm{\phi}_{v,H}$ and $\bm{\phi}_{a,H}$ from the hyperbolic to tangent space, using Eq.~(\ref{eqn:tangent_projection}).

The alignment loss plays a crucial role in guiding the network to learn informative modal representations. By using this loss, we hope to encourage the audio-visual features $\bm{\phi}_v$ and $\bm{\phi}_a$ to reflect more hierarchical structures. To achieve this, we transform these features into a curvature-negative space before performing the alignment operation. Additionally, since it is unclear which of the modalities contributes more to hierarchical exploration, our model ensures alignment between features by minimizing the difference in similarities within each modality. Motivated by the approach of \cite{tung2019similarity}, we design the loss using pairwise feature similarities within each of the mini-batch. The hyperbolic alignment loss is written as follows:
\begin{equation} \label{eqn:l_align}
\mathcal{L}_{align} = \frac{1}{N^2} \| \mathbf{M}_{v,norm} - \mathbf{M}_{a,norm} \|^2_F
\end{equation}
where $\mathbf{M}_{v, norm}, \mathbf{M}_{a, norm} \in \R^{N\times N}$ are the normalized matrices of intra-modal similarity matrices, $\mathbf{M}_v, \mathbf{M}_a \in \R^{N\times N}$, for a batch of video or audio features, $\bm{\phi}_{v,Tg}$ or $\bm{\phi}_{a,Tg}$, in the hyperbolic tangent space. Hence, we have $\mathbf{M}_{norm} = \frac{\mathbf{M}}{\| \mathbf{M} \|_F}$. The batch size is $N$ and $\| . \|_F$ calculates the Frobenius norm. 

Each element in $\mathbf{M}$ is computed via the cosine similarity between the $i^{\mathrm{th}}$ and $j^{\mathrm{th}}$ feature on the tangent space in a batch: $m_{i,j} = \text{cos}(\bm{\phi}_{Tg,i}, \bm{\phi}_{Tg,j}), i,j \in \{1, 2, ..., N \}$. Essentially, the loss $\mathcal{L}_{align}$ in Eq.~(\ref{eqn:l_align}) computes the mean element-wise squared difference between two similarity matrices. It aligns features from two modalities by minimizing the distance between two within-modal similarities. Finally, the overall loss of Hyper-alignment is comprised of two losses: 
\begin{equation} \label{eqn:l_total}
\mathcal{L}_{total} = \mathcal{L}_{avca} + \mathcal{L}_{align}
\end{equation}
where $\mathcal{L}_{avca}$ is the loss of the baseline. It is noted Hyper-alignment does not bring any extra weight to the network. 

\subsection{Hyper-single}
We propose Hyper-alignment to align features in a constant-curvature space. However, the representation ability of audio-visual features might suffer as the fixed curvature may not be suitable for the complex audio-visual data structure \cite{gao2022curvature}. To address this, we devise \textit{Hyper-single}, which utilizes an adaptive curvature. Unlike Hyper-alignment, which conforms to a fixed hyperbolic structure, Hyper-single produces a learnable curvature that enables flexible exploration of intrinsic data structures.

Particularly, as illustrated in Figure~\ref{fig:hypermodel} (b), the modality features, $\bm{\phi}_v$ and $\bm{\phi}_a$ are first concatenated into one vector $\bm{\phi}_{va} \in \R^{600}$, and $\bm{\phi}_{va}$ passes through one multiple-layer perceptron (MLP) to compute the curvature. Then, we have the curvature adaptively learned as follows:
\begin{equation}
    c = c_0 \cdot \mathrm{sigmoid}(\mathrm{MLP}(\bm{\phi}_{va}))
\end{equation}
where $\mathrm{MLP}(.): \R^{600} \rightarrow \R$ is one fully-connected layer and $c_0 < 0$ is the initial curvature. Hyper-single shares the same loss function in Eq.~(\ref{eqn:l_total}) as Hyper-alignment. Compared to Hyper-alignment, Hyper-single introduces the extra weights in $\mathrm{MLP}$ to the network.

\subsection{Hyper-multiple}
For the design of \textit{Hyper-multiple}, as depicted in Figure~\ref{fig:hypermodel} (c), we first attain several curvatures. Then, multiple Hyper-alignment modules are used for learning the alignment losses via the obtained curvatures. The multiple-curvature method leads to more generic spaces since diverse manifold structures may coexist in audio-visual data \cite{gao2021curvature}.

Hyper-multiple generates a set of adaptive curvatures $\bm{c} = \{ c_1, c_2, ..., c_{N_c} \}$ and synchronously map $\bm{\phi}_{v}$ and $\bm{\phi}_{a}$ into $N_c$ hyperbolic tangent spaces with these curvatures. Similar to Hyper-single, we make use of $\mathrm{MLP}(.): \R^{600} \rightarrow \R^{N_c}$ to output curvatures $\bm{c}$. The total loss in Eq.~(\ref{eqn:l_total}) applies Hyper-multiple, where the alignment loss $\mathcal{L}_{align}$ becomes:
\begin{equation}
    \mathcal{L}_{align} = \frac{1}{N_c}\sum_{i}^{N_c} \mathcal{L}_{align, i}
\end{equation}
where $\mathcal{L}_{align, i}$ is the hyperbolic alignment loss given in Eq.~(\ref{eqn:l_align}) with the $i^{\mathrm{th}}$ curvature, $i \in \{1, 2, ..., N_c\}$. Instead of restricting to only single curvature, Hyper-multiple computes multiple appropriate curvatures to learn more generic embeddings.

\section{Experiments}
In this section, we provide extensive experiments to verify the effectiveness of our method. The proposed \textit{Hyperbolic Alignment Module} is evaluated in audio-visual zero-shot classification on three datasets: VGGSound-GZSL, UCF-GZSL, and ActivityNet-GZSL. The details of each dataset are as follows.

\textbf{VGGSound-GZSL} \cite{mercea2022audio} is a modified version of the audio-visual dataset VGGSound \cite{chen2020vggsound}.
\textbf{UCF-GZSL} \cite{mercea2022audio} is a subset of the action video recognition dataset UCF101 \cite{soomro2012ucf101} that includes audio information. 
\textbf{ActivityNet-GZSL} \cite{mercea2022audio} is based on the action recognition dataset ActivityNet \cite{caba2015activitynet}. The statistics of the class split for three datasets are provided in Table~1 of \cite{mercea2022audio} and Table~2 of the supplementary material of \cite{mercea2022audio}.

\begin{table*}[th]
\centering
\resizebox{1.0\textwidth}{!}{
\begin{tabular}{l|cccc|cccc|cccc}
\hline
&\multicolumn{4}{c|}{\textbf{VGGSound-GZSL}} 
&\multicolumn{4}{c|}{\textbf{UCF-GZSL}} 
&\multicolumn{4}{c}{\textbf{ActivityNet-GZSL}}  \\

Method &$\mathrm{S}$ \textcolor{black}{$\uparrow$} &$\mathrm{U}$ \textcolor{black}{$\uparrow$} &$\mathrm{HM}$ \textcolor{black}{$\uparrow$} &$\mathrm{ZSL}$ \textcolor{black}{$\uparrow$} &$\mathrm{S}$ \textcolor{black}{$\uparrow$} &$\mathrm{U}$ \textcolor{black}{$\uparrow$} &$\mathrm{HM}$ \textcolor{black}{$\uparrow$} &$\mathrm{ZSL}$ \textcolor{black}{$\uparrow$} &$\mathrm{S}$ \textcolor{black}{$\uparrow$} &$\mathrm{U}$ \textcolor{black}{$\uparrow$} &$\mathrm{HM}$ \textcolor{black}{$\uparrow$} &$\mathrm{ZSL}$ \textcolor{black}{$\uparrow$} \\ \hline

CJME \cite{parida2020coordinated}      &8.69 &4.78 &6.17 &5.16   &26.04 &8.21 &12.48 &8.29   &5.55 &4.75 &5.12 &5.84 \\
AVGZSLNet \cite{mazumder2021avgzslnet} &18.05 &3.48 &5.83 &5.28   &52.52 &10.90 &18.05 &13.65   &8.93 &5.04 &6.44 &5.40 \\
AVCA \cite{mercea2022audio} &14.90 &4.00 &6.31 &6.00   &51.53 &18.43 &27.15 &20.01   &24.86 &8.02 &12.13 &9.13 \\ \hline

\rowcolor{gray} Hyper-alignment &13.22 &5.01 &7.27  &6.14    &57.28 &17.83 &27.19  &19.02   &23.50 &8.47 &12.46  &9.83  \\

\rowcolor{gray} Hyper-single    &9.79 &6.23 &7.62  &6.46    &52.67 &19.04 &27.97  &22.09    &23.60 &10.13 &\textbf{14.18}  &\textbf{10.80}  \\

\rowcolor{gray} Hyper-multiple  &15.02 &6.75 &\textbf{9.32}  &\textbf{7.97}    &63.08 &19.10 &\textbf{29.32}  &\textbf{22.24}    &23.38 &8.67 &12.65  &9.50  \\ \hline

\end{tabular}}
\vspace{1mm}
\caption{Experimental results of audio-visual zero-shot learning on three datasets (main feature). AVCA \cite{mercea2022audio} is adopted as the baseline for the proposed Hyper-alignment, Hyper-single, and Hyper-multiple modules. The best results in $\mathrm{HM}$ and $\mathrm{ZSL}$ are in \textbf{bold}. The curvatures of Hyper-alignment on VGGSound-GZSL, UCF-GZSL, and ActivityNet-GZSL are set as $-0.5$, $-0.2$, and $-0.2$. The numbers of adaptive curvatures $N_{c}$ for Hyper-multiple are $2$, $3$ and $2$.} \label{tab:main}
\end{table*}

\begin{table*}[t]
\centering
\resizebox{1.0\textwidth}{!}{
\begin{tabular}{l|cccc|cccc|cccc}
\hline
&\multicolumn{4}{c|}{\textbf{VGGSound-GZSL}$^{cls}$} 
&\multicolumn{4}{c|}{\textbf{UCF-GZSL}$^{cls}$} 
&\multicolumn{4}{c}{\textbf{ActivityNet-GZSL}$^{cls}$}  \\

Method &$\mathrm{S}$ \textcolor{black}{$\uparrow$} &$\mathrm{U}$ \textcolor{black}{$\uparrow$} &$\mathrm{HM}$ \textcolor{black}{$\uparrow$} &$\mathrm{ZSL}$ \textcolor{black}{$\uparrow$} &$\mathrm{S}$ \textcolor{black}{$\uparrow$} &$\mathrm{U}$ \textcolor{black}{$\uparrow$} &$\mathrm{HM}$ \textcolor{black}{$\uparrow$} &$\mathrm{ZSL}$ \textcolor{black}{$\uparrow$} &$\mathrm{S}$ \textcolor{black}{$\uparrow$} &$\mathrm{U}$ \textcolor{black}{$\uparrow$} &$\mathrm{HM}$ \textcolor{black}{$\uparrow$} &$\mathrm{ZSL}$ \textcolor{black}{$\uparrow$} \\ \hline

CJME \cite{parida2020coordinated}      &10.86 &2.22 &3.68 &3.72   &33.89 &24.82 &28.65 &29.01   &10.75 &5.55 &7.32 &6.29 \\
AVGZSLNet \cite{mazumder2021avgzslnet} &15.02 &3.19 &5.26 &4.81   &74.79 &24.15 &36.51 &31.51   &13.70 &5.96 &8.30 &6.39 \\
AVCA \cite{mercea2022audio}            &12.63 &6.19 &8.31 &6.91   &63.15 &30.72 &41.34 &37.72   &16.77 &7.04 &9.92 &7.58 \\ \hline

\rowcolor{gray} Hyper-alignment &12.50 &6.44 &8.50  &7.25    &57.13 &33.86 &42.52  &39.80    &29.77 &8.77 &13.55  &9.13  \\

\rowcolor{gray} Hyper-single    &12.56 &5.03 &7.18 &5.47   &63.47 &34.85 &44.99  &39.86    &24.61 &10.10 &14.32  &10.37  \\

\rowcolor{gray} Hyper-multiple  &15.62 &6.00 &\textbf{8.67}  &\textbf{7.31}    &74.26 &35.79 &\textbf{48.30}  &\textbf{52.11}     &36.98 &9.60 &\textbf{15.25}  &\textbf{10.39}  \\ \hline

\end{tabular}}
\vspace{1mm}
\caption{Experimental results of audio-visual zero-shot learning on three datasets (cls feature). AVCA \cite{mercea2022audio} is adopted as the baseline for the proposed Hyper-alignment, Hyper-single, and Hyper-multiple modules. The best results in $\mathrm{HM}$ and $\mathrm{ZSL}$ are in \textbf{bold}. The curvatures of Hyper-alignment on VGGSound-GZSL$^{cls}$, UCF-GZSL$^{cls}$, and ActivityNet-GZSL$^{cls}$ are set as $-0.1$, $-0.2$ and $-0.2$. The numbers of adaptive curvatures $N_{c}$ for Hyper-multiple are $3$, $2$ and $3$.} \label{tab:cls}
\end{table*}

We use classical metrics $\mathrm{S}$ and $\mathrm{U}$ for evaluating the performances of seen and unseen classes, respectively. We also use the metric $\mathrm{ZSL}$ on unseen classes. In addition to $\mathrm{ZSL}$, which solely evaluates the performance for unseen classes, the metric harmonic mean (HM) is used. It considers both seen and unseen classes' performances, $\mathrm{S}$ and $\mathrm{U}$, in the test stage: $\mathrm{HM} = \frac{2\mathrm{US}}{\mathrm{U+S}}$. 

All these metrics are evaluated on both the ``main feature" and ``cls feature," which indicate the pre-trained models used to extract data features $\bm{v}$, $\bm{a}$, and $\bm{w}$ \cite{mercea2022audio}. To fairly compare the performances of the Hyperbolic Alignment Module and the baseline, we keep all settings of hyperparameters in AVCA \cite{mercea2022audio}, including the learning rate, training epoch, \etc. In this case, $c_0$ is set as $-0.4$. In the ablation study, we analyze the effects of different parameters: curvature $c$ of Hyper-alignment and the number of curvatures $N_c$ of Hyper-multiple.

\subsection{Result Analysis}
The main results of the proposed modules for audio-visual zero-shot learning are presented in Table~\ref{tab:main} and \ref{tab:cls}. In general, Hyper-alignment outperforms the baseline in most cases. For instance, on UCF-GZSL$^{cls}$, Hyper-alignment achieves an accuracy of $42.52\%/39.80\%$ in $\mathrm{HM/ZSL}$, which is higher than the baseline's accuracy of $41.34\%/37.72\%$. Hyper-single, which computes an adaptive curvature for flexible exploration of audio-visual data structures, performs better than Hyper-alignment, except in the case of VGGSound-GZSL$^{cls}$. For example, on ActivityNet-GZSL, Hyper-single surpasses Hyper-alignment in $\mathrm{HM/ZSL}$ by $2.05\%/0.97\%$. Meanwhile, Hyper-multiple, which learns multiple adaptive curvatures, clearly outperforms Hyper-single, except in the case of ActivityNet-GZSL. 

As evidenced by our results on UCF-GZSL, Hyper-multiple outperforms all other methods, achieving a performance of $29.32\%/22.24\%$. Moreover, the classification accuracy for seen classes ($\mathrm{S}$) using Hyper-multiple is typically higher than those of other approaches. This suggests that the Hyper-multiple module generates more generic representations that are better suited for audio-visual zero-shot learning. In light of these results, we conclude that Hyper-multiple is the preferred design.

In conclusion, our proposed Hyperbolic Alignment Module improves audio-visual zero-shot learning performance over the baseline, as demonstrated in Table \ref{tab:main} and \ref{tab:cls}, with effective geometric alignments.

\subsection{Ablation Study}
\noindent \textbf{Curvature.} Experiments of Hyper-alignment are conducted under varying curvatures to understand how the degree to which the Euclidean space is distorted by hyperbolic geometry affects the model's performance. In addition to the hyperbolic space ($c<0$), feature alignments are also performed in Euclidean space ($c=0$) and spherical space ($c>0$). The results of these experiments are shown in Table~\ref{tab:curvature1}, which indicate that the performance of Hyper-alignment on UCF-GZSL$^{cls}$ and ActivityNet-GZSL$^{cls}$ peaks at a curvature of $-0.2$. This suggests that there is an optimal curvature for Hyper-alignment that achieves the best performance. Moreover, we observe that the model is robust against the curvature of the hyperbolic space, as performances with different curvatures are better than the baseline in most cases.

\begin{table}[t]
\centering
\resizebox{0.48\textwidth}{!}{
\begin{tabular}{l|c|cc|cc}
\hline
&
&\multicolumn{2}{c|}{\textbf{UCF-GZSL}$^{cls}$}  
&\multicolumn{2}{c}{\textbf{ActivityNet-GZSL}$^{cls}$} \\

Method  &$c$  &$\mathrm{HM}$ \textcolor{black}{$\uparrow$} &$\mathrm{ZSL}$ \textcolor{black}{$\uparrow$}   &$\mathrm{HM}$ \textcolor{black}{$\uparrow$} &$\mathrm{ZSL}$ \textcolor{black}{$\uparrow$} \\ \hline

AVCA \cite{mercea2022audio}      &-     &41.34 &37.72   &9.92 &7.58 \\ \hline

&1.0   &38.62 &31.64   &9.26  &6.23 \\
&0.8   &35.33 &31.58   &10.24 &7.01 \\
&0.6   &38.80 &38.62   &10.28 &7.01 \\
&0.4   &38.69 &38.22   &10.55 &7.58 \\
&0.2   &38.56 &34.45   &10.56 &6.95 \\
\multirow{-6}{*}{Sphere-alignment}  
                  &0.1   &34.14 &31.72   &10.13 &6.74 \\ \hline

Euclidean-alignment  &0.0   &38.82 &36.01   &10.47 &8.00 \\ \hline

\rowcolor{gray}                      &-0.1  &40.64 &38.09   &12.51 &8.67 \\
\rowcolor{gray}                      &-0.2  &\textbf{42.52} &\textbf{39.80}   &\textbf{13.55} &9.13 \\
\rowcolor{gray}                      &-0.4  &41.62 &36.83   &13.15 &9.08 \\ 
\rowcolor{gray}                      &-0.6  &37.69 &36.56   &12.86 &\textbf{9.77} \\ 
\rowcolor{gray}                      &-0.8  &35.48 &36.16   &12.63 &9.62 \\ 
\rowcolor{gray} \multirow{-6}{*}{Hyper-alignment}   
                                     &-1.0  &38.25 &34.11   &10.95 &7.30 \\ \hline

\end{tabular}}
\vspace{1mm}
\caption{Ablation study: curvature. Different curvature values of Hyper-alignment are tested on UCF-GZSL$^{cls}$ and ActivityNet-GZSL$^{cls}$. Besides Hyper-alignment, Euclidean-alignment and Sphere-alignment are also considered. From the table, Hyper-alignment shows advantages over Euclidean-alignment and Sphere-alignment.} \label{tab:curvature1}
\end{table}

\noindent \textbf{Mixed curvature learning.} In this experiment, variants of multiple geometric curvatures are evaluated. Hyper-multiple combines two hyperbolic alignments when the number of curvatures $N_c$ is set to $2$. Other combinations, such as Euclidean and spherical alignments, spherical and hyperbolic alignments, \etc, are also tested. The results in Table~\ref{tab:curvature2} show that the combination of two hyperbolic alignments ($\mathrm{H+H}$) achieves the best performance among all the combinations. This observation is consistent with the results in Table~\ref{tab:curvature1}, which demonstrate the advantage of hyperbolic geometry over other geometries in audio-visual zero-shot learning.

\begin{table}[t]
\centering
\resizebox{0.48\textwidth}{!}{
\begin{tabular}{l|c|cc|cc}
\hline
&
&\multicolumn{2}{c|}{\textbf{UCF-GZSL}$^{cls}$}  
&\multicolumn{2}{c}{\textbf{ActivityNet-GZSL}$^{cls}$} \\

Method  &$c$  &$\mathrm{HM}$ \textcolor{black}{$\uparrow$} &$\mathrm{ZSL}$ \textcolor{black}{$\uparrow$}   &$\mathrm{HM}$ \textcolor{black}{$\uparrow$} &$\mathrm{ZSL}$ \textcolor{black}{$\uparrow$} \\ \hline

AVCA \cite{mercea2022audio}      &-      &41.34 &37.72   &9.92 &7.58 \\ \hline
\rowcolor{gray}                      &$\mathrm{E+E}$  &39.26 &37.05   &9.83  &7.35 \\
\rowcolor{gray}                      &$\mathrm{E+S}$  &38.31 &36.71   &11.93 &8.03 \\
\rowcolor{gray}                      &$\mathrm{E+H}$  &42.73 &37.74   &7.49  &5.70 \\
\rowcolor{gray}                      &$\mathrm{S+S}$  &23.97 &33.02   &10.02 &6.32 \\
\rowcolor{gray}                      &$\mathrm{S+H}$  &40.37 &38.16   &12.25 &7.78 \\
\rowcolor{gray} \multirow{-6}{*}{Mixed-curvature learning}                                      
                                     &$\mathrm{H+H}$  &\textbf{48.30} &\textbf{52.11} &\textbf{13.70} &\textbf{8.95}  \\ \hline
\end{tabular}}
\vspace{1mm}
\caption{Ablation study: mixed curvature learning. Different variants of mixed-curvature geometries are tested on UCF-GZSL$^{cls}$ and ActivityNet-GZSL$^{cls}$. $\mathrm{H}$, $\mathrm{S}$ and $\mathrm{E}$ indicate hyperbolic, spherical and Euclidean alignments. From the table, Geometry-multiple, which incorporates two hyperbolic geometries, gains the most performance improvements.} \label{tab:curvature2}
\end{table}

\noindent \textbf{Effectiveness of multiple curvatures.} Here, we test the performance of Hyper-multiple with different numbers of adaptive curvatures, $N_c$, on VGGSound-GZSL and VGGSound-GZSL$^{cls}$. The results in Table~\ref{tab:curvature_num} demonstrate that Hyper-multiple with $2$ or $3$ adaptive curvatures achieves the best performance. This suggests that, for Hyper-multiple, having more curvatures does not necessarily lead to better performance in audio-visual zero-shot learning.

\begin{table}[t]
\centering
\resizebox{0.48\textwidth}{!}{
\begin{tabular}{l|c|cc|cc}
\hline
&
&\multicolumn{2}{c|}{\textbf{VGGSound-GZSL}}  
&\multicolumn{2}{c}{\textbf{VGGSound-GZSL}$^{cls}$} \\

Method  &$N_c$  &$\mathrm{HM}$ \textcolor{black}{$\uparrow$} &$\mathrm{ZSL}$ \textcolor{black}{$\uparrow$}  &$\mathrm{HM}$ \textcolor{black}{$\uparrow$} &$\mathrm{ZSL}$ \textcolor{black}{$\uparrow$} \\ \hline

AVCA \cite{mercea2022audio}      &-      &6.31 &6.00   &8.31 &6.91 \\ \hline
\rowcolor{gray}                      &1  &7.62 &6.46   &7.18 &5.47 \\
\rowcolor{gray}                      &2  &\textbf{9.32} &\textbf{7.97}   &\textbf{8.97} &6.72\\
\rowcolor{gray} \multirow{-3}{*}{Hyper-multiple}                     
                                     &3  &8.91 &7.12   &8.67 &\textbf{7.31} \\ \hline

\end{tabular}}
\vspace{1mm}
\caption{Ablation study: effectiveness of multiple curvatures. Different curvature numbers of Hyper-multiple are evaluated on VGGSound-GZSL and VGGSound-GZSL$^{cls}$.} \label{tab:curvature_num}
\end{table}

\noindent \textbf{Audio-visual feature alignment.} In the main paper, we present the hyperbolic alignment modules that align the visual and audio features $\bm{\phi}_v$ and $\bm{\phi}_a$ before the cross-attention module of the baseline. In this experiment, we investigate aligning the features after the cross-attention and compare the results in Table~\ref{tab:location}. As seen from the table, aligning $\bm{\phi}_v$ and $\bm{\phi}_a$ before the cross-attention achieves the best result. On the other hand, aligning $\bm{\phi}_{v,att}$ and $\bm{\phi}_{a,att}$ harms the performance. These results suggest that the cross-modal fusion performed by the cross-attention module may alter the hierarchy of the data's original features from each modality.

\begin{table}[t]
\centering
\resizebox{0.48\textwidth}{!}{
\begin{tabular}{l|c|cc|cc}
\hline
&
&\multicolumn{2}{c|}{\textbf{UCF-GZSL}}  
&\multicolumn{2}{c}{\textbf{ActivityNet-GZSL}} \\

Method  &aligned feature  &$\mathrm{HM}$ \textcolor{black}{$\uparrow$} &$\mathrm{ZSL}$ \textcolor{black}{$\uparrow$}  &$\mathrm{HM}$ \textcolor{black}{$\uparrow$} &$\mathrm{ZSL}$ \textcolor{black}{$\uparrow$} \\ \hline

AVCA \cite{mercea2022audio}      &-             &27.15 &20.01   &12.13 &9.13 \\ \hline
\rowcolor{gray}                  &$\bm{\phi}_v/\bm{\phi}_a$              &\textbf{27.97} &\textbf{22.09}   &\textbf{14.18} &\textbf{10.80} \\
\rowcolor{gray}                  &$\bm{\phi}_{v,att}/\bm{\phi}_{a,att}$   &13.78 &11.89   &8.71 &6.59 \\ 
\rowcolor{gray} \multirow{-3}{*}{Hyper-single}                                      
                                 &$\bm{\theta}_{v}/\bm{\theta}_{a}$       &25.58 &18.77   &12.96 &10.00 \\\hline
\end{tabular}}
\vspace{1mm}
\caption{Ablation study: audio-visual feature alignment. Different locations of aligned features by Hyper-single are tested on datasets UCF-GZSL and ActivityNet-GZSL.} \label{tab:location}
\end{table}

\subsection{Discussion}
\begin{figure*}[h]
\centering
	\includegraphics[width=1.0\linewidth]{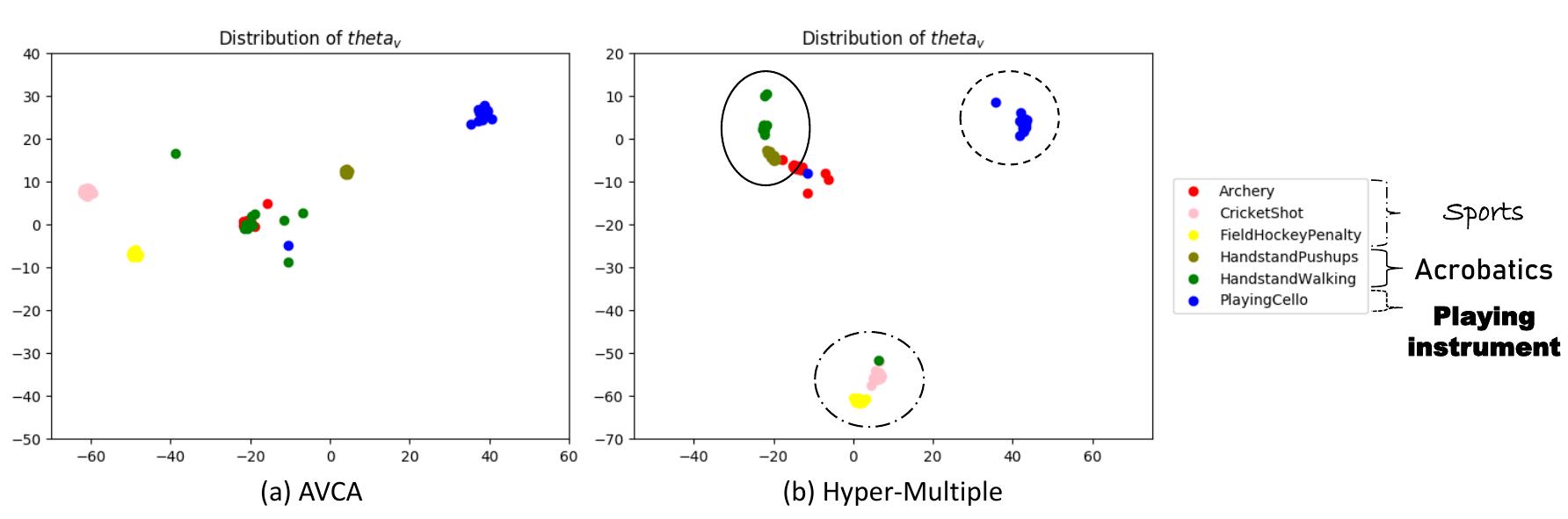}
 	\caption{Visualization examples on UCF-GZSL$^{cls}$. We give t-SNE visualizations of $\bm{\theta}_v$ from six seen classes: ``Archery'', ``CricketShot'', ``FieldHockeyPenalty'', ``HandstandPushups'', ``HandstandWalking'' and ``PlayingCello''. They can be categorized into three superclasses: ``Sports'', ``Acrobatics'' and ``Playinginstruments''. Hyper-multiple, which learns a hyperbolic alignment loss, pushes away features from different superclasses.}
 	\label{fig:visual1}
\end{figure*}

\begin{figure}[h]
\centering
	\includegraphics[width=1.0\linewidth]{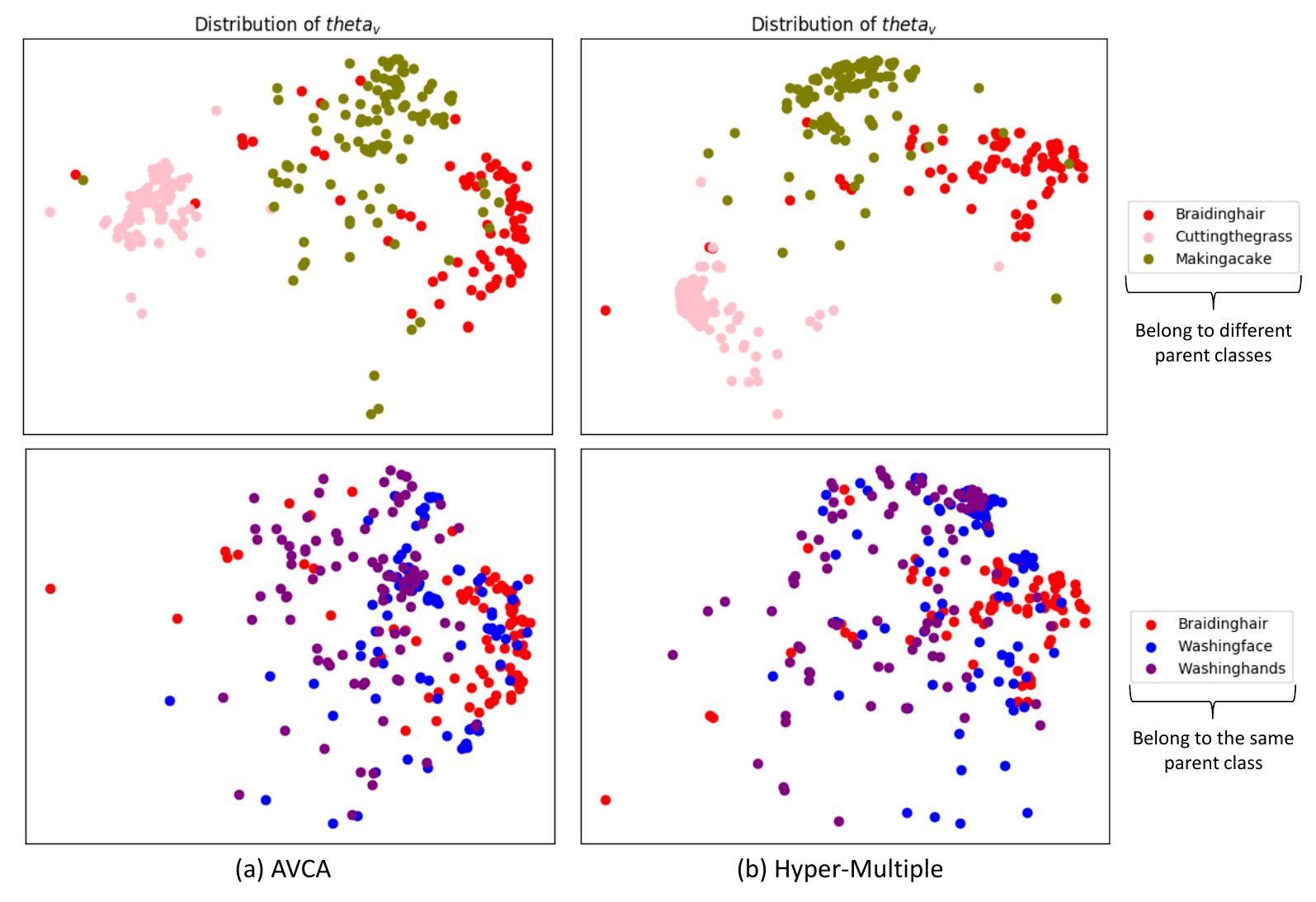}
 	\caption{Visualization examples on ActivityNet-GZSL$^{cls}$. We give t-SNE visualizations of $\bm{\theta}_v$. The features $\bm{\theta}_v$ of three unseen classes, which are from different parent classes, are visualized in the upper figure. The lower figure depicts the feature distribution of three unseen classes from the same parent class. Comparing two figures, Hyper-multiple pushes away features more from different superclasses than the same superclass.}
 	\label{fig:visual2}
\end{figure}

\begin{table}[b]
\centering
\resizebox{0.48\textwidth}{!}{
\begin{tabular}{l|c|c}
\hline
&\textbf{VGGSound-GZSL}
&\textbf{ActivityNet-GZSL}$^{cls}$ \\ \hline

Method    &$\delta_{rel}$ &$\delta_{rel}$ \\ \hline

AVCA \cite{mercea2022audio}            &0.30  &0.37    \\
Hyper-alignment                        &0.27  &0.32    \\
\hline
\end{tabular}}
\vspace{1mm}
\caption{Discussion: $\delta$-Hyperbolicity. We compute $\delta_{rel}$ on the feature $\bm{\theta}_v$. The smaller value of $\delta_{rel}$ indicates the stronger hyperbolicity inside $\bm{\theta}_v$. Hyper-alignment gets a lower value of $\delta_{rel}$ than ACVA, which indicates the representations via the hyperbolic alignment loss learn more hierarchical structures.} \label{tab:hierarchy}
\end{table}

\noindent \textbf{$\delta$-Hyperbolicity.} 
The underlying structures of audio-visual datasets may exhibit high non-Euclidean latent geometry since there is a rich hierarchy in audio-visual data. In this part, we attempt to demonstrate that the proposed hyperbolic alignment modules are capable of uncovering some hierarchical knowledge. To measure the degree of the hierarchy inside the features, we use the $\delta$-Hyperbolicity metric, which has been shown to be effective in detecting hierarchical structure \cite{khrulkov2020hyperbolic, fournier2015computing}. In our case, we compute $\delta_{rel}$ \cite{khrulkov2020hyperbolic, fournier2015computing}, a metric overall testing sample features that indicate the degree to which the feature distribution resembles a tree-like structure. Smaller values of $\delta_{rel}$ reflect stronger hyperbolicity within the feature space. As we can see from Table~\ref{tab:hierarchy}, Hyper-alignment achieves lower $\delta_{rel}$ values than AVCA on VGGSound-GZSL and ActivityNet-GZSL$^{cls}$. The results support our hypothesis that aligned representations using hyperbolic alignment loss convey more hierarchical information.

\noindent \textbf{t-SNE Visualizations.} 
In addition to measuring $\delta$-Hyperbolicity, we provide t-SNE visualizations that reveal the capability of hyperbolic spaces in exploring data hierarchy. As shown in Figure~\ref{fig:visual1}, the hyperbolic alignment loss facilitates pulling together features from the same parent class while pushing away features belonging to different parent classes on UCF-GZSL$^{cls}$. For instance, features of "CricketShot" and "FieldHockeyPenalty", which belong to the same parent class "Sports," are pulled together, while features of "CricketShot" and "HandstandWalking", which belong to different parent classes, are pushed away. To some extent, this explains how hyperbolic alignment between visual and audio features in hyperbolic spaces enables more accurate audio-visual zero-shot classification. Similarly, the visualization in Figure \ref{fig:visual2} suggests that on ActivityNet$^{cls}$, features belonging to different superclasses are more discriminately distributed than features from the same superclass. These visualizations somewhat describe how our approach uncovers the hierarchical knowledge among classes.

\section{Conclusion}
We introduce a novel hyperbolic alignment module that improves cross-modality representations in audio-visual zero-shot learning. This is achieved by leveraging the properties of hyperbolic spaces to explore more hierarchical structures within the audio-visual data and generate more distinctive features. Our approach includes three modules, Hyper-alignment, Hyper-single, and Hyper-multiple, for computing the loss, and we demonstrate their effectiveness through empirical evaluation. Our use of curvature-aware geometric learning to leverage data hierarchy may inspire further research in the field of audio-visual learning.

\section*{Acknowledgements}
We acknowledge the support from the Australian Research Council (ARC) for M. Harandi's project DP230101176 and the Southeast University Start-Up Grant for New Faculty (No. 4009002309). This research work is also supported by the Big Data Computing Center of Southeast University.

{\small
\bibliographystyle{ieee_fullname}
\bibliography{egbib}
}

\end{document}